\documentclass[10pt,twocolumn,letterpaper]{article}

\usepackage{cvpr}
\usepackage{times}
\usepackage{epsfig}
\usepackage{graphicx}
\usepackage{amsmath}
\usepackage{amssymb}
\usepackage{booktabs}
\usepackage{multirow}
\usepackage[dvipsnames]{xcolor}

\usepackage{bbm}
\usepackage{enumitem}
\usepackage{gensymb}
\usepackage{svg}
\usepackage[font=small]{caption}
\usepackage{lipsum}%

\makeatletter
\renewcommand{\paragraph}{%
  \@startsection{paragraph}{4}%
  {\z@}{1.5ex \@plus 1ex \@minus .2ex}{-1em}%
  {\normalfont\normalsize\bfseries}%
}

\usepackage[pagebackref=true,breaklinks=true,letterpaper=true,colorlinks,bookmarks=false]{hyperref}

\cvprfinalcopy %

\def\cvprPaperID{5346} %

\ifcvprfinal\fi
\begin{document}

\setlength{\skip\footins}{10pt}

\title{Disentangling and Unifying Graph Convolutions\\for Skeleton-Based Action Recognition}

\author{
Ziyu Liu\textsuperscript{1,3}, Hongwen Zhang\textsuperscript{2}, Zhenghao Chen\textsuperscript{1}, Zhiyong Wang\textsuperscript{1}, Wanli Ouyang\textsuperscript{1,3}\\
{\textsuperscript{1}The University of Sydney\quad \textsuperscript{2}University of Chinese Academy of Sciences \& CASIA}\\
{\textsuperscript{3}The University of Sydney, SenseTime Computer Vision Research Group, Australia}
\\
}

\maketitle
\thispagestyle{empty}

\begin{abstract}
Spatial-temporal graphs have been widely used by skeleton-based action recognition algorithms to model human action dynamics.
To capture robust movement patterns from these graphs, long-range and multi-scale context aggregation and spatial-temporal dependency modeling are critical aspects of a powerful feature extractor.
However, existing methods have limitations in achieving (1) unbiased long-range joint relationship modeling under multi-scale operators and (2) unobstructed cross-spacetime information flow for capturing complex spatial-temporal dependencies.
In this work, we present (1) a simple method to disentangle multi-scale graph convolutions and (2) a unified spatial-temporal graph convolutional operator named G3D.
The proposed multi-scale aggregation scheme disentangles the importance of nodes in different neighborhoods for effective long-range modeling.
The proposed G3D module leverages dense cross-spacetime edges as skip connections for direct information propagation across the spatial-temporal graph.
By coupling these proposals, we develop a powerful feature extractor named MS-G3D based on which our model\footnote{Code is available at \href{https://github.com/kenziyuliu/ms-g3d}{\tt github.com/kenziyuliu/ms-g3d}} outperforms previous state-of-the-art methods on three large-scale datasets: NTU RGB+D 60, NTU RGB+D 120, and Kinetics Skeleton 400.
\end{abstract}

\section{Introduction} \label{sec:intro}

Human action recognition is an important task with many real-world applications.
In particular, \textit{skeleton-based} human action recognition involves predicting actions from skeleton representations of human bodies instead of raw RGB videos, and the significant results seen in recent work~\cite{ST-GCN, 2s-AGCN, dgnn, attention-gcn-lstm-cvpr2019, AS-GCN-skeleton-cvpr19, li2018co-human-skeleton-ijcai18, view-adaptive-human-skeleton-iccv2017, spatial-reasoning-tempora-stack-human-skeleton} have proven its merits.
In contrast to RGB representations, skeleton data contain only the 2D \cite{ST-GCN, kinetics-dataset} or 3D \cite{ntu60, ntu-rgbd-120} positions of the human key joints, providing highly abstract information that is also free of environmental noises (\eg background clutter, lighting conditions, clothing), allowing action recognition algorithms to focus on the robust features of the action.

\begin{figure}[t]
\centering
\includegraphics[width=\linewidth]{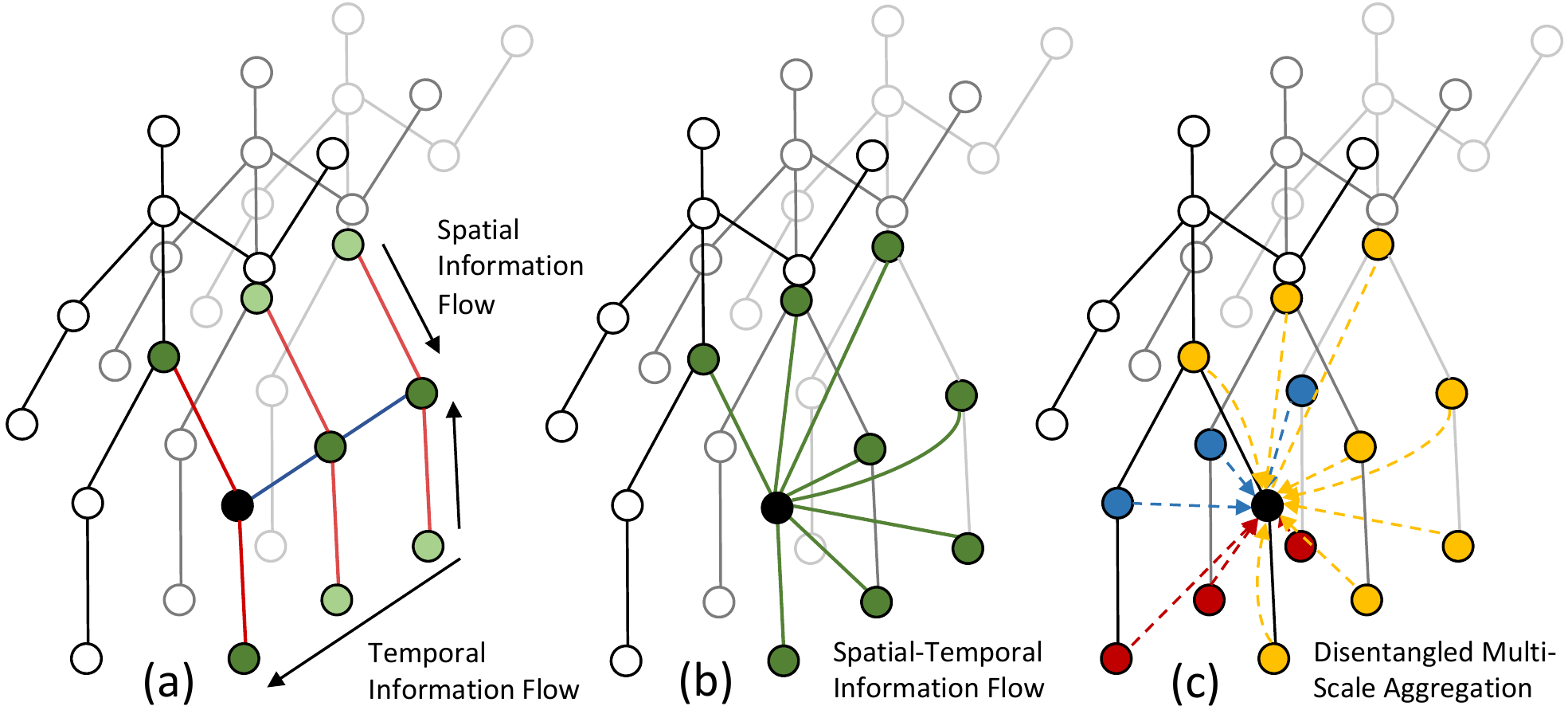}
\caption{
   (a) Factorized spatial and temporal modeling on skeleton graph sequences causes \textit{indirect} information flow.
   (b) In this work, we propose to capture cross-spacetime correlations with \textit{unified} spatial-temporal graph convolutions.
   (c) Disentangling node features at separate spatial-temporal neighborhoods (\textcolor{Dandelion}{yellow}, \textcolor{NavyBlue}{blue}, \textcolor{BrickRed}{red} at different distances, partially colored for clarity) is pivotal for effective multi-scale learning in the spatial-temporal domain.
}
\label{fig:intro-figure}
\vspace{-3mm}
\end{figure}

Earlier approaches to skeleton-based action recognition treat human joints as a set of independent features, and they model the spatial and temporal joint correlations through hand-crafted \cite{lie-group-human-skeleton, wang2012mining} or learned \cite{ntu60, hierarchical-rnn-human-skeleton, memory-attention-net-human-skeleton, view-adaptive-human-skeleton-iccv2017} aggregations of these features.
However, these methods overlook the inherent relationships between the human joints, which are best captured with human skeleton graphs with joints as nodes and their natural connectivity (\ie ``bones'') as edges.
For this reason, recent approaches \cite{ST-GCN, st-gcn-concur-18, attention-gcn-lstm-cvpr2019, spatial-reasoning-tempora-stack-human-skeleton, dgnn} model the joint movement patterns of an action with a skeleton \textit{spatial-temporal graph}, which is a series of disjoint and isomorphic skeleton graphs at different time steps carrying information in both spatial and temporal dimensions.

For robust action recognition from skeleton graphs, an ideal algorithm should look beyond the local joint connectivity and extract multi-scale structural features and long-range dependencies, since joints that are structurally apart can also have strong correlations.
Many existing approaches achieve this by performing \textit{graph convolutions} \cite{GCN} with higher-order polynomials of the skeleton adjacency matrix: intuitively, a powered adjacency matrix captures the number of walks between every pair of nodes with the length of the walks being the same as the power; the adjacency polynomial thus increases the receptive field of graph convolutions by making distant neighbors reachable.
However, this formulation suffers from the \textit{biased weighting problem}, where the existence of \textit{cyclic} walks on undirected graphs means that edge weights will be biased towards closer nodes against further nodes. On skeleton graphs, this means that a higher polynomial order is only marginally effective at capturing information from distant joints, since the aggregated features will be dominated by the joints from local body parts.
This is a critical drawback limiting the scalability of existing multi-scale aggregators.

Another desirable characteristic of robust algorithms is the ability to leverage the complex cross-spacetime joint relationships for action recognition.
However, to this end, most existing approaches \cite{ST-GCN, 2s-AGCN, st-gcn-concur-18, dgnn, AS-GCN-skeleton-cvpr19, attention-gcn-lstm-cvpr2019, st-graph-routing-skeleton-aaai19} deploy interleaving spatial-only and temporal-only modules (Fig.~\ref{fig:intro-figure}(a)), analogous to factorized 3D convolutions \cite{P3D, tran2018closer}.
A typical approach is to first use graph convolutions to extract spatial relationships at each time step, and then use recurrent \cite{st-gcn-concur-18, attention-gcn-lstm-cvpr2019, st-graph-routing-skeleton-aaai19} or 1D convolutional \cite{ST-GCN, 2s-AGCN, AS-GCN-skeleton-cvpr19, dgnn} layers to model temporal dynamics.
While such factorization allows efficient long-range modeling, it hinders the direct information flow across spacetime for capturing complex regional spatial-temporal joint dependencies.
For example, the action ``standing up'' often has co-occurring movements of upper and lower body across both space and time, where upper body movements (leaning forward) strongly correlate to the lower body's \textit{future} movements (standing up).
These strong cues for making predictions may be ineffectively captured by factorized modeling.

In this work, we address the above limitations from two aspects.
First, we propose a new multi-scale aggregation scheme that tackles the biased weighting problem by removing redundant dependencies between further and closer neighborhoods, thus disentangling their features under multi-scale aggregation (illustrated in Fig.~\ref{fig:exp-vs-dis-multi-scale}). This leads to more powerful multi-scale operators that can model relationships of joints irrespective of the distances between them.
Second, we propose G3D, a novel unified spatial-temporal graph convolution module that \textit{directly} models cross-spacetime joint dependencies.
G3D does so by introducing graph edges across the ``3D'' spatial-temporal domain as skip connections for unobstructed information flow~(Fig.~\ref{fig:intro-figure}(b)), substantially facilitating spatial-temporal feature learning.
Remarkably, our proposed disentangled aggregation scheme augments G3D with multi-scale reasoning in spacetime (Fig.~\ref{fig:intro-figure}(c)) without being affected by the biased weighting problem, despite extra edges were introduced.
The resulting powerful feature extractor, named MS-G3D, forms a building block of our final model architecture that outperforms state-of-the-art methods on three large-scale skeleton action datasets:
NTU RGB+D 120 \cite{ntu-rgbd-120}, NTU RGB+D 60 \cite{ntu60}, and Kinetics Skeleton 400 \cite{kinetics-dataset}.
The main contributions of this work are summarized as follows:
\\ [2mm]
(i) We propose a disentangled multi-scale aggregation scheme that removes redundant dependencies between node features from different neighborhoods, which allows powerful multi-scale aggregators to effectively capture graph-wide joint relationships on human skeletons.
\\ [2mm]
(ii) We propose a unified spatial-temporal graph convolution (G3D) operator which facilitates direct information flow across spacetime for effective feature learning.
\\ [2mm]
(iii)
Integrating the disentangled aggregation scheme with G3D gives a powerful feature extractor (MS-G3D) with multi-scale receptive fields across both spatial and temporal dimensions.
The direct multi-scale aggregation of features in spacetime further boosts model performance.

\section{Related Work}

\subsection{Neural Nets on Graphs}

\paragraph{Architectures.}
To extract features from arbitrarily structured graphs, Graph Neural Networks (GNNs) have been developed and explored extensively \cite{chebynet, GCN, spectral-cnn-yann-lecun, diffusion-cnn, graph-sage, graph-attention-networks,GIN-how-powerful-are-gnns,mixhop,graph-unet,hammond-spectral-2011,adaptive-gcn}.
Recently proposed GNNs can broadly be classified into spectral GNNs \cite{spectral-cnn-yann-lecun, hammond-spectral-2011, adaptive-gcn,spectral-cnn-yann-lecun-2,GCN} and spatial GNNs \cite{GCN,GIN-how-powerful-are-gnns,graph-sage,diffpool,deep-graph-infomax,mixhop,simplifying-gcns}.
Spectral GNNs convolve the input graph signals with a set of learned filters in the graph Fourier domain.
They are however limited in terms of computational efficiency and generalizability to new graphs due to the requirement of eigendecomposition and the assumption of fixed adjacency.
Spatial GNNs, in contrast, generally perform layer-wise update for each node by (1) selecting neighbors with a neighborhood function (\eg adjacent nodes); (2) merging the features from the selected neighbors and itself with an aggregation function (\eg mean pooling); and (3) applying an activated transformation to the merged features (\eg MLP \cite{GIN-how-powerful-are-gnns}). Among different GNN variants, the Graph Convolutional Network (GCN) \cite{GCN} was first introduced as a first-order approximation for localized spectral convolutions, but its simplicity as a mean neighborhood aggregator \cite{GIN-how-powerful-are-gnns, graph-survey-2019} has quickly led many subsequent spatial GNN architectures~\cite{GIN-how-powerful-are-gnns, mixhop, simplifying-gcns, graph-unet} and various applications involving graph structured data \cite{video-as-spacetime-region-graphs, graph-wavenet, stgcn-traffic-2017, ST-GCN, 2s-AGCN, attention-gcn-lstm-cvpr2019, AS-GCN-skeleton-cvpr19} to treat it as a \textit{spatial} GNN baseline. This work adapts the layer-wise update rule in GCN.

\paragraph{Multi-Scale Graph Convolutions.}
Multi-scale spatial GNNs have also been proposed to capture features from non-local neighbors.
\cite{mixhop, st-gcn-concur-18, AS-GCN-skeleton-cvpr19, simplifying-gcns, lanczos-net} use higher order polynomials of the graph adjacency matrix to aggregate features from long-range neighbor nodes.
Truncated Block Krylov network \cite{stronger-gcns-krylov} similarly raises the adjacency matrix to higher powers and obtains multi-scale information through dense features concatenation from different hidden layers.
LanczosNet \cite{lanczos-net} deploys a low-rank approximation of the adjacency matrix to speed up the exponentiation on large graphs.
As mentioned in Section~\ref{sec:intro}, we argue that adjacency powering can have adverse effects on long-range modeling due to weighting bias, and our proposed module aims to address this with disentangled multi-scale aggregators.

\subsection{Skeleton-Based Action Recognition}

Earlier approaches \cite{lie-group-human-skeleton, hierarchical-rnn-human-skeleton, ntu60, song2017end, wang2012mining, memory-attention-net-human-skeleton, view-adaptive-human-skeleton-iccv2017} to skeleton-based action recognition focus on hand-crafting features and joint relationships for downstream classifiers, which ignore the important semantic connectivity of the human body. By constructing spatial-temporal graphs and modeling the spatial relationships with GNNs directly, recent approaches \cite{ST-GCN, st-gcn-concur-18, GR-GCN,AS-GCN-skeleton-cvpr19,GR-GCN,2s-AGCN,dgnn,attention-gcn-lstm-cvpr2019,st-graph-routing-skeleton-aaai19} have seen significant performance boost, indicating the necessity of the semantic human skeleton for action predictions.

An early application of graph convolutions is ST-GCN \cite{ST-GCN}, where spatial graph convolutions along with interleaving temporal convolutions are used for spatial-temporal modeling.
A concurrent work by Li et al.~\cite{st-gcn-concur-18} presents a similar approach, but it notably introduces a multi-scale module by raising skeleton adjacency to higher powers. AS-GCN \cite{AS-GCN-skeleton-cvpr19} also uses adjacency powering for multi-scale modeling, but it additionally generates human poses to augment the spatial graph convolution. Spatial-Temporal Graph Routing (STGR) network \cite{st-graph-routing-skeleton-aaai19} adds extra edges to the skeleton graph using frame-wise attention and global self-attention mechanisms. Similarly, 2s-AGCN \cite{2s-AGCN} introduces graph adaptiveness with self-attention along with a freely learned graph residual mask. It also uses a two-stream ensemble with skeleton bone features to boost performance. DGNN \cite{dgnn} likewise leverages bone features, but it instead simultaneously updates the joint and bone features through an alternating spatial aggregation scheme. Note that these approaches primarily focus on spatial modeling; in contrast, we present a unified approach for capturing complex joint correlations directly across spacetime.

Another relevant work is GR-GCN \cite{GR-GCN}, which merges every three frames over the skeleton graph sequence and adds sparsified edges between adjacent frames. Whereas GR-GCN also deploys cross-spacetime edges, our G3D module has several important distinctions:
(1) Cross-spacetime edges in G3D follow the semantic human skeleton, which is naturally a more interpretable and more robust representation than the sparsified, one-size-fits-all graph in GR-GCN. The underlying graph is also much easier to compute.
(2) GR-GCN has cross-spacetime edges only between adjacent frames, which prevents it to reason beyond a limited temporal context of three frames.
(3) G3D can learn from multiple temporal contexts simultaneously leveraging different window sizes and dilations, which is not addressed in GR-GCN.

\section{MS-G3D}

\subsection{Preliminaries}

\paragraph{Notations.}

A human skeleton graph is denoted as $\mathcal{G} = (\mathcal{V}, \mathcal{E})$, where $\mathcal{V} = \{v_1, ..., v_N\}$ is the set of $N$ nodes representing joints, and $\mathcal{E}$ is the edge set representing bones captured by an adjacency matrix $\mathbf{A} \in \mathbb{R}^{N \times N}$ where initially $\mathbf{A}_{i,j} = 1$ if an edge directs from $v_i$ to $v_j$ and 0 otherwise.
$\mathbf{A}$ is symmetric since $\mathcal{G}$ is undirected.
Actions as \textit{graph sequences} have a node features set $\mathcal{X} = \{x_{t,n} \in \mathbb{R}^{C} \mid t,n \in \mathbb{Z}, 1\leq t\leq T, 1\leq n \leq N\}$ represented as a feature tensor $\mathbf{X} \in \mathbb{R}^{T \times N \times C}$, where $x_{t,n} = \mathbf{X}_{t,n,:}$ is the $C$ dimensional feature vector for node $v_n$ at time $t$ over a total of $T$ frames.
The input action is thus adequately described by $\mathbf{A}$ structurally and by $\mathbf{X}$ feature-wise, with $\mathbf{X}_t \in \mathbb{R}^{N\times C}$ being the node features at time $t$. $\Theta^{(l)} \in \mathbb{R}^{C_l \times C_{l+1}}$ denotes a learnable weight matrix at layer $l$ of a network.

\paragraph{Graph Convolutional Nets (GCNs).}

On skeleton inputs defined by features $\mathbf{X}$ and graph structure $\mathbf{A}$, the layer-wise update rule of GCNs can be applied to features at time $t$ as:
\begin{equation} \label{eq:gcn}
    \mathbf{X}_t^{(l+1)} = \sigma\left(
        \tilde{\mathbf{D}}^{-\frac{1}{2}}
        \tilde{\mathbf{A}}
        \tilde{\mathbf{D}}^{-\frac{1}{2}}
        \mathbf{X}_t^{(l)}
        \Theta^{(l)}
    \right),
\end{equation}
where $\tilde{\mathbf{A}} = \mathbf{A + I}$ is the skeleton graph with added \textit{self-loops} to keep identity features, $\tilde{\mathbf{D}}$ is the diagonal degree matrix of $\tilde{\mathbf{A}}$, and $\sigma(\cdot)$ is an activation function.
The term
$
\tilde{\mathbf{D}}^{-\frac{1}{2}}
\tilde{\mathbf{A}}
\tilde{\mathbf{D}}^{-\frac{1}{2}}
\mathbf{X}_t^{(l)}
$
can be intuitively interpreted as an approximate spatial \textit{mean} feature aggregation from the direct neighborhood followed by an activated linear layer.

\subsection{Disentangled Multi-Scale Aggregation}

\paragraph{Biased Weighting Problem.} \label{sec:biased-weighting-problem}

\begin{figure}[t]
\centering
\includegraphics[width=\linewidth]{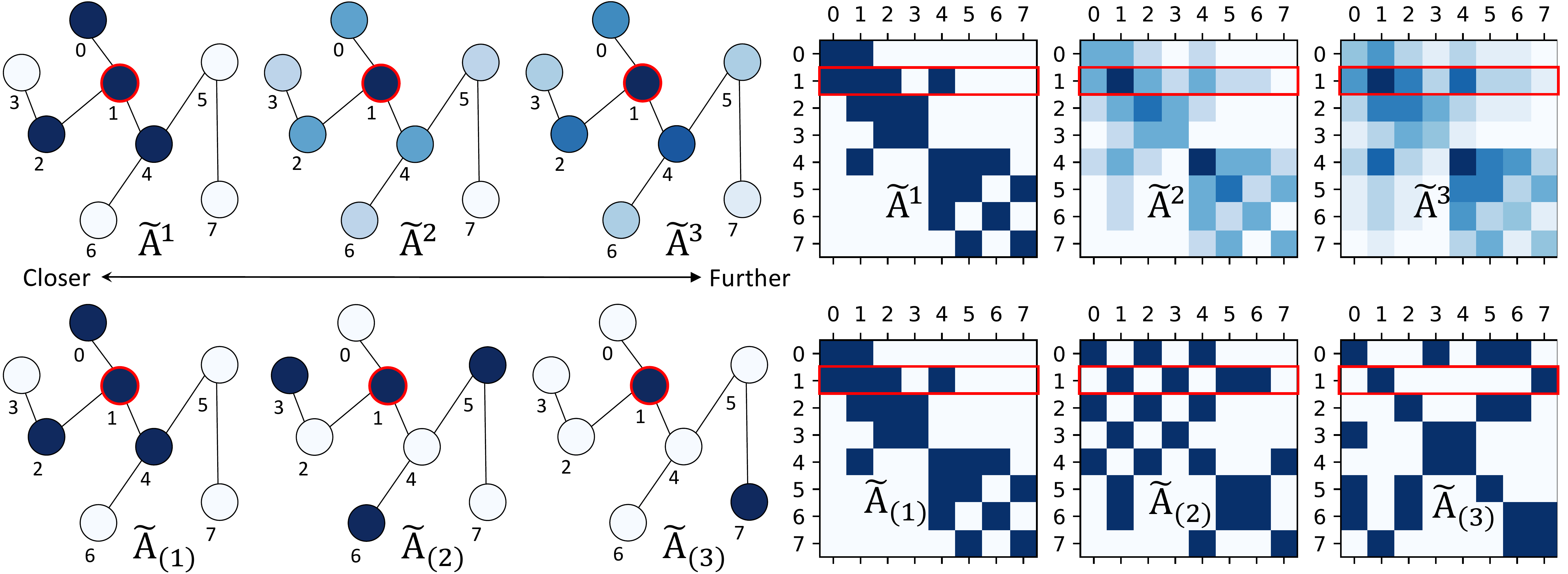}
   \caption{Illustration of the \textbf{biased weighting problem} and the proposed \textbf{disentangled aggregation scheme}. Darker color indicates higher weighting to the central node (\textcolor{BrickRed}{red}). \textbf{Top left}: closer nodes receive higher weighting from adjacency powering, which makes long-range modeling less effective, especially when multiple scales are aggregated. \textbf{Bottom left}: our proposed disentangled aggregation models joint relationships at each neighborhood while keeping identity features. \textbf{Right}: Visualizing the corresponding adjacency matrices. Node self-loops are omitted for visual clarity.}
\label{fig:exp-vs-dis-multi-scale}
\end{figure}

Under the spatial aggregation framework in Eq.~\ref{eq:gcn},
existing approaches~\cite{AS-GCN-skeleton-cvpr19} employ \textit{higher-order polynomials} of the adjacency matrix to aggregate multi-scale structural information at time $t$, as:
\begin{equation} \label{eq:existing-multi-scale}
    \mathbf{X}_t^{(l+1)} = \sigma \left(
        \sum_{k=0}^{K}
            \widehat{\mathbf{A}}^{k}
            \mathbf{X}_t^{(l)}
            \Theta^{(l)}_{(k)}
    \right),
\end{equation}
where $K$ controls the number of scales to aggregate. Here, $\widehat{\mathbf{A}}$ is a normalized form of $\mathbf{A}$, \eg \cite{st-gcn-concur-18} uses the symmetric normalized graph Laplacian $\widehat{\mathbf{A}} = \mathbf{L}^{\text {norm}} = \mathbf{I}-\mathbf{D}^{-\frac{1}{2}} \mathbf{A} \mathbf{D}^{-\frac{1}{2}}$; \cite{AS-GCN-skeleton-cvpr19} uses the random-walk normalized adjacency $\widehat{\mathbf{A}}=\mathbf{D}^{-1} \mathbf{A}$; more generally, one can use $\widehat{\mathbf{A}}=\tilde{\mathbf{D}}^{-\frac{1}{2}}\tilde{\mathbf{A}}\tilde{\mathbf{D}}^{-\frac{1}{2}}$ from GCNs.
It is easy to see that $\mathbf{A}^k_{i,j}=\mathbf{A}^k_{j,i}$ gives the number of length $k$ walks between $v_i$ and $v_j$, and thus the term $\widehat{\mathbf{A}}^k\mathbf{X}_t^{(l)}$ is performing a \textit{weighted} feature average based on the number of such walks.
However, it is clear that there are drastically more possible length $k$ walks to closer nodes than to the actual $k$-hop neighbors due to cyclic walks. This causes a bias towards the local region as well as nodes with higher degrees. The node self-loops in GCNs allow even more possible cycles (as walks can always cycle on self-loops) and thus amplify the bias. See Fig.~\ref{fig:exp-vs-dis-multi-scale} for illustration.
Under multi-scale aggregation on skeleton graphs, the aggregated features will thus be dominated by signals from local body parts, making it ineffective to capture long-range joint dependencies with higher polynomial orders.

\paragraph{Disentangling Neighborhoods.} \label{sec:disentangling-neighborhoods}

To address the above problem, we first define the $k$-adjacency matrix $\tilde{\mathbf{A}}_{(k)}$ as
\begin{equation} \label{eq:k-adjacency}
    [\tilde{\mathbf{A}}_{(k)}]_{i,j} = \left\{
        \begin{array}{ll}
            {1} & {\text { if } d(v_i, v_j) = k }, \\
            {1} & {\text { if } i = j }, \\
            {0} & {\text { otherwise, }}
        \end{array}\right.
\end{equation}
where $d(v_i, v_j)$ gives the shortest distance in number of hops between $v_i$ and $v_j$. $\tilde{\mathbf{A}}_{(k)}$ is thus a generalization of $\tilde{\mathbf{A}}$ to further neighborhoods, with $\tilde{\mathbf{A}}_{(1)} = \tilde{\mathbf{A}}$ and $\tilde{\mathbf{A}}_{(0)} = \mathbf{I}$.
Under spatial aggregation in Eq.~\ref{eq:gcn}, the inclusion of self-loops in $\tilde{\mathbf{A}}_{(k)}$ is critical for learning the \textit{relationships} between the current joint and its $k$-hop neighbors, as well as for keeping each joint's \textit{identity} information when no $k$-hop neighbors are available. Given that $N$ is small, $\tilde{\mathbf{A}}_{(k)}$ can be easily computed,
\eg, using \textit{differences} of graph powers as
$\tilde{\mathbf{A}}_{(k)} = \mathbf{I} + \mathbbm{1}\left(\tilde{\mathbf{A}}^k \geq 1\right) - \mathbbm{1}\left(\tilde{\mathbf{A}}^{k-1} \geq 1\right)$.
Substituting $\widehat{\mathbf{A}}^{k}$ with $\tilde{\mathbf{A}}_{(k)}$ in Eq.~\ref{eq:existing-multi-scale}, we arrive at:
\begin{equation} \label{eq:disentangle-multi-scale}
    \mathbf{X}_t^{(l+1)} = \sigma \left(
            \sum_{k=0}^{K}
                \tilde{\mathbf{D}}_{(k)}^{-\frac{1}{2}}
                \tilde{\mathbf{A}}_{(k)}
                \tilde{\mathbf{D}}_{(k)}^{-\frac{1}{2}}
                \mathbf{X}_t^{(l)}
                \Theta^{(l)}_{(k)}
        \right),
\end{equation}
where $\tilde{\mathbf{D}}_{(k)}^{-\frac{1}{2}}\tilde{\mathbf{A}}_{(k)}\tilde{\mathbf{D}}_{(k)}^{-\frac{1}{2}}$ is the normalized \cite{GCN} $k$-adjacency.

Unlike the previous case where possible length $k$ walks are predominantly conditioned on length $k-1$ walks, the proposed disentangled formulation in Eq.~\ref{eq:disentangle-multi-scale} addresses the biased weighting problem by removing redundant dependencies of distant neighborhoods' weighting on closer neighborhoods.
Additional scales with larger $k$ are therefore aggregated in an \textit{additive} manner under a multi-scale operator, making long-range modeling with large values of $k$ to remain effective.
The resulting $k$-adjacency matrices
are also more sparse than their exponentiated counterparts (see Fig.~\ref{fig:exp-vs-dis-multi-scale}), allowing more efficient representations.

\subsection{G3D: Unified Spatial-Temporal Modeling}

Most existing work treats skeleton actions as a sequence of disjoint graphs where features are extracted through spatial-only (\eg GCNs) and temporal-only (\eg TCNs) modules.
We argue that such factorized formulation is less effective for capturing complex spatial-temporal joint relationships.
Clearly, if a strong connection exists between a pair of nodes, then during layer-wise propagation the pair should incorporate a significant portion each other's features to reflect such a connection \cite{ST-GCN, 2s-AGCN, attention-gcn-lstm-cvpr2019}. However, as signals are propagated across spacetime through a series of local aggregators (GCNs and TCNs alike), they are weakened as redundant information is aggregated from an increasingly larger spatial-temporal receptive field. The problem is more evident if one observes that GCNs do not perform a weighted aggregation to distinguish each neighbor.

\paragraph{Cross-Spacetime Skip Connections.}

To tackle the above problem, we propose a more reasonable approach to allow cross-spacetime \textit{skip connections}, which are readily modeled with cross-spacetime \textit{edges} in a spatial-temporal graph.
Let us first consider a sliding temporal window of size $\tau$ over the input graph sequence, which, at each step, obtains a spatial-temporal subgraph $\mathcal{G}_{(\tau)} = (\mathcal{V}_{(\tau)}, \mathcal{E}_{(\tau)})$ where
$\mathcal{V}_{(\tau)} = \mathcal{V}_1 \cup ... \cup \mathcal{V}_\tau$ is the union of all node sets across $\tau$ frames in the window.
The initial edge set $\mathcal{E}_{(\tau)}$ is defined by tiling $\tilde{\mathbf{A}}$ into a block adjacency matrix $\tilde{\mathbf{A}}_{(\tau)}$, where
\begin{equation} \label{eq:g3d-graph-definition}
    \tilde{\mathbf{A}}_{(\tau)}
    = \left[
        \begin{array}{ccc}
            {\tilde{\mathbf{A}}} & {\cdots} & {\tilde{\mathbf{A}}} \\
            {\vdots} & {\ddots} & {\vdots} \\
            {\tilde{\mathbf{A}}} & {\cdots} & {\tilde{\mathbf{A}}}
        \end{array}
    \right] \in \mathbb{R}^{\tau N \times \tau N}.
\end{equation}
Intuitively, each submatrix $[\tilde{\mathbf{A}}_{(\tau)}]_{i,j} = \tilde{\mathbf{A}}$ means every node in $\mathcal{V}_i$ is connected to itself and its 1-hop spatial neighbors at frame $j$ by \textit{extrapolating} the frame-wise spatial connectivity (which is $[\tilde{\mathbf{A}}_{(\tau)}]_{i,i}$ for all $i$) to the temporal domain.
Thus, each node within $\mathcal{G}_{(\tau)}$ is densely connected to itself and its 1-hop spatial neighbors across all $\tau$ frames.
We can easily obtain $\mathbf{X}_{(\tau)} \in \mathbb{R}^{T\times \tau N \times C}$ using the same sliding window over $\mathbf{X}$ with zero padding to construct $T$ windows.
Using Eq.~\ref{eq:gcn}, we thus arrive at a \textit{unified} spatial-temporal graph convolutional operator for the $t^\text{th}$ temporal window:
\begin{equation} \label{eq:g3d-update-rule}
    [\mathbf{X}_{(\tau)}^{(l+1)}]_t = \sigma\left(
        \tilde{\mathbf{D}}_{(\tau)}^{-\frac{1}{2}}
        \tilde{\mathbf{A}}_{(\tau)}
        \tilde{\mathbf{D}}_{(\tau)}^{-\frac{1}{2}}
        [\mathbf{X}_{(\tau)}^{(l)}]_t
        \Theta^{(l)}
    \right).
\end{equation}

\paragraph{Dilated Windows.}

Another significant aspect of the above window construction is that the frames need not to be adjacent. A \textit{dilated} window with $\tau$ frames and a dilation rate $d$ can be constructed by picking a frame every $d$ frames, and reusing the same spatial-temporal structure $\tilde{\mathbf{A}}_{(\tau)}$. Similarly, we can obtain node features $\mathbf{X}_{(\tau, d)} \in \mathbb{R}^{T\times \tau N \times C}$
($d=1$ if omitted) and perform layer-wise update as in Eq.~\ref{eq:g3d-update-rule}. Dilated windows allow larger temporal receptive fields \textit{without} growing the size of $\tilde{\mathbf{A}}_{(\tau)}$, analogous to how dilated convolutions \cite{dilated-conv} keep constant complexities.

\paragraph{Multi-Scale G3D.}
We can also integrate the proposed disentangled multi-scale aggregation scheme (Eq.~\ref{eq:disentangle-multi-scale}) into G3D for multi-scale reasoning \textit{directly} in the spatial-temporal domain.
We thus derive the MS-G3D module from Eq.~\ref{eq:g3d-update-rule} as:
\begin{equation} \label{eq:g3d-multi-scale}
\small [\mathbf{X}_{(\tau)}^{(l+1)}]_t =
    \sigma \left(
    \sum_{k=0}^{K}
        \tilde{\mathbf{D}}_{(\tau, k)}^{-\frac{1}{2}}
        \tilde{\mathbf{A}}_{(\tau, k)}
        \tilde{\mathbf{D}}_{(\tau, k)}^{-\frac{1}{2}}
        [\mathbf{X}_{(\tau)}^{(l)}]_t
        \Theta^{(l)}_{(k)}
    \right),
\end{equation}
where $\tilde{\mathbf{A}}_{(\tau, k)}$ and $\tilde{\mathbf{D}}_{(\tau, k)}$ are defined similarly as $\tilde{\mathbf{A}}_{(k)}$ and $\tilde{\mathbf{D}}_{(k)}$ respectively.
Remarkably, our proposed disentangled aggregation scheme complements this unified operator, as G3D's increased node degrees from spatial-temporal connectivity can contribute to the biased weighting problem.

\paragraph{Discussion.}
We give more in-depth analyses on G3D as follows.
(1) It is analogous to classical 3D convolutional blocks \cite{3d-conv-original}, with its spatial-temporal receptive field defined by $\tau$, $d$, and $\tilde{\mathbf{A}}$.
(2) Unlike 3D convolutions, G3D's parameter count from $\Theta_{(\cdot)}^{(\cdot)}$ is independent of $\tau$ or $|\mathcal{E}_{(\tau)}|$, making it generally less prone to overfitting with large $\tau$.
(3) The dense cross-spacetime connections in G3D entail a trade-off on $\tau$, as larger values of $\tau$ bring larger temporal receptive fields at the cost of more \textit{generic} features due to larger immediate neighborhoods. Additionally, larger $\tau$ implies a quadratically larger $\tilde{\mathbf{A}}_{(\tau)}$ and thus more operations with multi-scale aggregation.
On the other hand, larger dilations $d$ bring larger temporal coverage at the cost of temporal \textit{resolution} (lower frame rates). $\tau$ and $d$ thus must be balanced carefully.
(4) G3D modules are designed to capture complex \textit{regional} spatial-temporal instead of long-range dependencies that are otherwise more economically captured by factorized modules. We thus observe the best performance when G3D modules are augmented with long-range, factorized modules, which we discuss in the next section.

\subsection{Model Architecture}

\begin{figure*}[h]
\includegraphics[width=\linewidth]{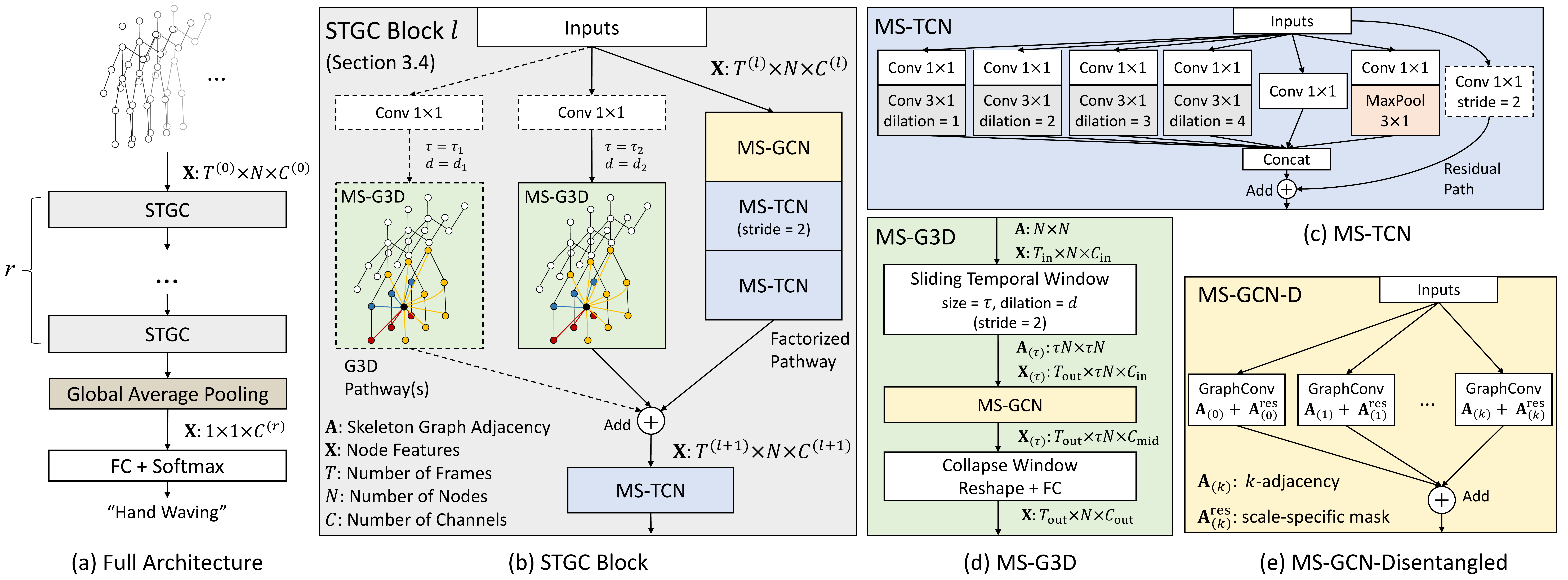}
   \caption{(Match components with colors) \textbf{Architecture Overview}. ``TCN'', ``GCN'', prefix ``MS-'', and suffix ``-D'' denotes temporal and graph convolutional blocks, and multi-scale and disentangled aggregation, respectively (Section~\ref{sec:disentangling-neighborhoods}). Each of the $r$ STGC blocks (b) deploys a multi-pathway design to capture long-range and regional spatial-temporal dependencies simultaneously.
   \textbf{Dotted modules}, including extra G3D pathway, 1$\times$1 conv, and strided temporal convolutions, are situational for model performance/complexity trade-off.
    }
\label{fig:architecture}
\end{figure*}

\paragraph{Overall Architecture.}
The final model architecture is illustrated in Fig.~\ref{fig:architecture}.
On a high level, it contains a stack of $r$ spatial-temporal graph convolutional (STGC) blocks to extract features from skeleton sequences, followed by a global average pooling layer and a softmax classifier.
Each STGC block deploys two types of pathways to simultaneously capture complex regional spatial-temporal joint correlations as well as long-range spatial and temporal dependencies:
(1) The G3D pathway first constructs spatial-temporal windows, performs disentangled multi-scale graph convolutions on them, and then \textit{collapses} them with a fully connected layer for window feature readout.
The extra dotted G3D pathway (Fig.~\ref{fig:architecture}(b)) indicates the model can learn from multiple spatial-temporal contexts concurrently with different $\tau$ and $d$;
(2) The factorized pathway augments the G3D pathway with long-range, spatial-only, and temporal-only modules: the first layer is a multi-scale graph convolutional layer capable of modeling the entire skeleton graph with the maximum $K$; it is then followed by two multi-scale temporal convolutions layers to capture extended temporal contexts (discussed below).
The outputs from all pathways are aggregated as the STGC block output, which has 96, 192, and 384 feature channels respectively within a typical $r$=3 block architecture.
Batch normalization \cite{batchnorm} and ReLU is added at the end of each layer except for the last layer. All STGC blocks, except the first, downsample the temporal dimension with stride 2 temporal conv and sliding windows.

\paragraph{Multi-Scale Temporal Modeling.}

The spatial-temporal windows $\mathcal{G}_{(\tau)}$ used by G3D are a closed structure by themselves, which means G3D must be accompanied by temporal modules for cross-window information exchange.
Many existing work \cite{ST-GCN,st-graph-routing-skeleton-aaai19,2s-AGCN,dgnn,AS-GCN-skeleton-cvpr19} performs temporal modeling using temporal convolutions with a fixed kernel size $k_t \times 1$ throughout the architecture. As a natural extension to our multi-scale spatial aggregation, we enhance vanilla temporal convolutional layers with multi-scale learning, as illustrated in Fig.~\ref{fig:architecture}(c).
To lower the computational costs due to the extra branches, we deploy a bottleneck design \cite{inception-v4}, fix kernel sizes at 3$\times$1, and use different dilation rates \cite{dilated-conv} instead of larger kernels for larger receptive fields. We also use residual connections \cite{resnet} to facilitate training.

\paragraph{Adaptive Graphs.}
To improve the flexibility of graph convolutional layers which performs homogeneous neighborhood averaging, we add a simple learnable, unconstrained graph residual mask $\mathbf{A}^\text{res}$ inspired by \cite{2s-AGCN,dgnn} to every $\tilde{\mathbf{A}}_{(k)}$ and $\tilde{\mathbf{A}}_{(\tau, k)}$ to strengthen, weaken, add, or remove edges dynamically. For example, Eq.~\ref{eq:disentangle-multi-scale} is updated to
\begin{equation} \label{eq:disentangle-multi-scale-adaptive}
    \small \mathbf{X}_t^{(l+1)} =
            \sigma \left(
            \sum_{k=0}^{K}
                \tilde{\mathbf{D}}_{(k)}^{-\frac{1}{2}}
                (\tilde{\mathbf{A}}_{(k)} + \mathbf{A}^\text{res}_{(k)})
                \tilde{\mathbf{D}}_{(k)}^{-\frac{1}{2}}
                \mathbf{X}_t^{(l)}
                \Theta^{(l)}_{(k)}
        \right).
\end{equation}
$\mathbf{A}^\text{res}$ is initialized with random values around zero and is different for each $k$ and $\tau$, allowing each multi-scale context (either spatial or spatial-temporal) to select the best suited mask. Note also that since $\mathbf{A}^\text{res}$ is optimized for all possible actions, which may have different optimal edge sets for feature propagation, it is expected to give minor edge corrections and may be insufficient when the graph structures have major deficiencies. In particular, $\mathbf{A}^\text{res}$ only partially mitigates the biased weighting problem (see Section~\ref{sec:component-studies}).

\paragraph{Joint-Bone Two-Stream Fusion.}

Inspired by the two-stream methods in \cite{2s-AGCN, dgnn, attention-gcn-lstm-cvpr2019} and the intuition that visualizing \textit{bones} along with joints can help humans recognize skeleton actions, we use a two-stream framework where a \textit{separate} model with identical architecture is trained using the bone features initialized as vector differences of adjacent joints directed away from the body center. The softmax scores from the joint/bone models are summed to obtain final prediction scores.
Since skeleton graphs are \textit{trees}, we add a zero bone vector at the body center to obtain $N$ bones from $N$ joints and reuse $\mathbf{A}$ for connectivity definition.

\section{Experiments} \label{sec:experiments}

\subsection{Datasets}

\paragraph{NTU RGB+D 60 and NTU RGB+D 120.}

NTU RGB+D 60 \cite{ntu60} is a large-scale action recognition dataset containing 56,578 skeleton sequences over 60 action classes captured from 40 distinct subjects and 3 different camera view angles.
Each skeleton graph contains $N=25$ body joints as nodes, with their 3D locations in space as initial features. Each frame of the action contains 1 to 2 subjects. The authors recommend reporting the classification accuracy under two settings: (1) Cross-Subject (X-Sub), where the 40 subjects are split into training and testing groups, yielding 40,091 and 16,487 training and testing examples respectively. (2) Cross-View (X-View), where all 18,932 samples collected from camera 1 are used for testing and the rest 37,646 samples used for training. NTU RGB+D 120 \cite{ntu-rgbd-120} extends NTU RGB+D 60 with an additional 57,367 skeleton sequences over 60 extra action classes, totalling 113,945 samples over 120 classes captured from 106 distinct subjects and 32 different camera setups.
The authors now recommend replacing the Cross-View setting with a Cross-Setup (X-Set) setting, where 54,468 samples collected from half of the camera setups are used for training and the rest 59,477 samples for testing. In Cross-Subject, 63,026 samples from a selected group of 53 subjects are used for training, and the rest 50,919 samples for testing.

\paragraph{Kinetics Skeleton 400.}

The Kinetics Skeleton 400 dataset is adapted from the Kinetics 400 video dataset \cite{kinetics-dataset} using the OpenPose \cite{openpose} pose estimation toolbox. It contains 240,436 training and 19,796 testing skeleton sequences over 400 classes, where each skeleton graph contains 18 body joints, along with their 2D spatial coordinates and the prediction confidence score from OpenPose as the initial joint features \cite{ST-GCN}. At each time step, the number of skeletons is capped at 2, and skeletons with lower overall confidence scores are discarded. Following the convention from \cite{kinetics-dataset, ST-GCN}, Top-1 and Top-5 accuracies are reported.

\subsection{Implementation Details}

Unless otherwise stated, all models have $r=3$ and are trained with SGD with momentum 0.9, batch size 32 (16 per worker), an initial learning rate 0.05 (can linearly scale up with batch size \cite{linearscalelr})  for 50, 60, and 65 epochs with step LR decay with a factor of 0.1 at epochs \{30, 40\}, \{30, 50\}, and \{45, 55\} for NTU RGB+D 60, 120, and Kinetics Skeleton 400, respectively. Weight decay is set to 0.0005 for final models and is adjusted accordingly during component studies. All skeleton sequences are padded to $T = 300$ frames by replaying the actions. Inputs are preprocessed with normalization and translation following \cite{2s-AGCN, dgnn}. No data augmentation is used for fair performance comparison.

\subsection{Component Studies} \label{sec:component-studies}
We analyze the individual components and their configurations in the final architecture. Unless stated, performance is reported as classification accuracy on the Cross-Subject setting of NTU RGB+D 60 using only the joint data.

\paragraph{Disentangled Multi-Scale Aggregation.}

We first justify our proposed disentangled multi-scale aggregation scheme by verifying its effectiveness with different number of scales over sparse and dense graphs. In Table \ref{tab:ablation-disentangled}, we do so using the individual pathways of the STGC blocks (Fig.~\ref{fig:architecture}(b)), referred to as ``GCN'' and ``G3D'', respectively, with suffixes ``-E'' and ``-D'' denoting adjacency powering and disentangled aggregation.
Here, the maximum $K=12$ is the diamater of skeleton graphs from NTU RGB+D 60, and we set $\tau=5$ for G3D modules.
To keep consistent normalization, we set $\widehat{\mathbf{A}} = \tilde{\mathbf{D}}^{-\frac{1}{2}} \tilde{\mathbf{A}} \tilde{\mathbf{D}}^{-\frac{1}{2}}$ in Eq.~\ref{eq:existing-multi-scale} for GCN-E and G3D-E.
We first observe that the disentangled formulation can bring as much as 1.4\% gain over simple adjacency powering at $K=4$, underpinning the necessity for neighborhood disentanglement.
In this case, the residual mask $\mathbf{A}^\text{res}$ partially corrects the weighting imbalance, narrowing the largest gap to 0.4\%.
However, the same set of experiments on the G3D pathway, where the window graph $\mathcal{G}_{(\tau)}$ is denser than the spatial graph $\mathcal{G}$, shows wider accuracy gaps between G3D-E and G3D-D, indicating a more severe biased weighting problem. In particular, we see 0.8\% performance gap at $K=12$ even if residual masks are added.
These results verify the effectiveness of the proposed disentangled aggregation scheme for multi-scale learning; it boosts performance across different number scales not only in the spatial domain, but more so in the spatial-temporal domain where it complements the proposed G3D module.
In general, the spatial GCNs benefits more from large $K$ than do the spatial-temporal G3D modules; for final architectures, we empirically set $K \in \{12,5\}$ for MS-GCN and MS-G3D blocks respectively.

\paragraph{Effectiveness of G3D.}

To validate the efficacy of G3D modules to capture complex spatial-temporal features, we build up the model incrementally with its individual components, and show its performance in Table \ref{tab:g3d-ablations}.
We use the joint stream from 2s-AGCN \cite{2s-AGCN} as the baseline for controlled experiments, and for fair comparison, we replaced its regular temporal convolutional layers with MS-TCN layers and obtained an improvement with less parameters.
First, we observe that the factorized pathway alone can outperform the baseline due to the powerful disentangled aggregation in MS-GCN.
However, if we simply scale up the factorized pathway to larger capacity (deeper and wider), or duplicate the factorized pathway to learn from different feature subspaces and mimic the multi-pathway design in STGC blocks, we observe limited gains.
In contrast, when the G3D pathway is added, we observe consistently better results with similar or less parameters, verifying G3D's ability to pick up complex regional spatial-temporal correlations that are previously overlooked by modeling spatial and temporal dependencies in a factorized fashion.

\begin{table}[t]
\centering
\begin{tabular}{lcccc}
\hline
\multirow{2}{*}{\textbf{Methods}} & \multicolumn{4}{c}{\textbf{Number of Scales}} \\
& \small $K=1$ & \small $K=4$ & \small $K=8$ & \small $K=12$ \\ \hline \hline
GCN-E & 85.1 & 85.6 & 86.5 & 86.6 \\
\textbf{GCN-D} & 85.1 & 87.0 & 86.9 & 86.8 \\ \hline
GCN-E + Mask & 86.1 & 87.0 & 87.5 & 87.7 \\
\textbf{GCN-D} + Mask & 86.1 & 86.9 & 87.9 & 87.8 \\ \hline \hline
G3D-E & 85.1 & 85.5 & 85.4 & 85.5 \\
\textbf{G3D-D} & 85.1 & 86.4 & 86.5 & 86.4 \\ \hline
G3D-E + Mask & 86.6 & 87.0 & 86.5 & 86.2 \\
\textbf{G3D-D} + Mask & 86.6 & 87.4 & 87.1 & 87.0 \\ \hline
\end{tabular}
\caption{
Accuracy (\%) with multi-scale aggregation on individual pathways of STGC blocks with different $K$.
``Mask'' refers to the residual masks $\mathbf{A}^\text{res}$.
If $K$\textgreater 1, GCN/G3D is \textbf{Multi-Scale} (\textbf{MS-}).
}
\label{tab:ablation-disentangled}
\vspace{-0mm}%
\end{table}

\begin{table}[t]
\centering
\begin{tabular}{lcc}
\hline
\textbf{Model Configurations} & \textbf{Params} & \textbf{Acc (\%)} \\
\hline
\hline
Baseline (Js-AGCN \cite{2s-AGCN}) & 3.5M & 86.0 \\
Baseline + MS-TCN & 1.6M & 86.7 \\
\hline
\small{MS-GCN (Factorized Pathway) Only} & 1.4M & 87.8 \\
\quad with $2.5\times$ Capacity & 3.5M & 88.5 \\
\quad with Dual Pathway & 2.8M & 88.6 \\
\hline
\hline
MS-GCN (Factorized Pathway) &  &  \\
\quad with MS-G3D ($\tau=3,d=1$) & 2.7M & 89.0 \\
\quad with MS-G3D ($\tau=3,d=2$) & 2.7M & 89.1 \\
\quad with MS-G3D ($\tau=3,d=3$) & 2.7M & 89.1 \\
\quad with MS-G3D ($\tau=5,d=1$) & 3.2M & 89.2 \\
\quad with MS-G3D ($\tau=5,d=2$) & 3.2M & 89.2 \\
\quad with MS-G3D ($\tau=7,d=1$)\textsuperscript{\dag} & 3.0M & 89.0 \\
\hline
\quad with 2 MS-G3D Pathways\textsuperscript{\dag} & \multirow{2}{*}{2.8M} &\multirow{2}{*}{89.3} \\
\quad\qquad $\tau = (3,3), d = (1,2)$ &  \\
\quad with 2 MS-G3D Pathways\textsuperscript{\dag} & \multirow{2}{*}{3.2M} &\multirow{2}{*}{89.4} \\
\quad\qquad $\tau = (3,5), d = (1,1)$ & & \\
\hline
\end{tabular}
\caption{
Model accuracy with various settings. MS-GCN and MS-G3D uses $K \in \{12, 5\}$ respectively.
\textsuperscript{\dag}Output channels double at the collapse window layer (Fig.~\ref{fig:architecture}(d), $C_\text{mid}$ to $C_\text{out}$) instead of at the graph convolution ($C_\text{in}$ to $C_\text{mid}$) to maintain similar budget.
}
\label{tab:g3d-ablations}
\vspace{-0mm}%
\end{table}

\begin{table}[t]
\centering
\begin{tabular}{lcc}
\hline
\textbf{G3D Graph Connectivity} & \textbf{Params} & \textbf{Acc (\%)} \\
\hline
\hline
(1) Grid-like & 2.7M & 88.7 \\
(2) Grid-like + dense self-edges & 2.7M & 88.6 \\
\hline
(Eq.~\ref{eq:g3d-graph-definition}) Cross-spacetime edges & 2.7M & 89.1 \\
\hline
\end{tabular}
\caption{Comparing graph connectivity settings ($\tau = 3, d = 2$).
}
\label{tab:ablation-graph-connectivity}
\vspace{-0mm}%
\end{table}

\begin{table}[t]
\centering
\begin{tabular}{lcc}
\hline
\multirow{2}{*}{\textbf{Methods}} & \multicolumn{2}{c}{\textbf{NTU RGB+D 120}} \\
 & X-Sub (\%) & X-Set (\%) \\
\hline
\hline
ST-LSTM \cite{st-lstm-trust-gates-skeleton-eccv2016} & 55.7 & 57.9 \\
GCA-LSTM \cite{two-stream-attention-lstm-ntu120} & 61.2 & 63.3 \\
RotClips + MTCNN \cite{rotclips-ntu120} & 62.2 & 61.8 \\
Body Pose Evolution Map \cite{body-pose-evolution-map-ntu-120} & 64.6 & 66.9 \\
\hline
2s-AGCN \cite{2s-AGCN} & 82.9 & 84.9 \\ \hline
\textbf{MS-G3D Net} & \textbf{86.9} & \textbf{88.4} \\
\hline
\end{tabular}
\caption{Classification accuracy comparison against state-of-the-art methods on the NTU RGB+D 120 Skeleton dataset.
}
\label{tab:ntu-120-results}
\vspace{-0mm}%
\end{table}

\begin{table}[t]
\centering
\begin{tabular}{lcc}
\hline
\multirow{2}{*}{\textbf{Methods}} & \multicolumn{2}{c}{\textbf{NTU RGB+D 60}} \\
 & X-Sub (\%) & X-View (\%) \\
\hline
\hline
IndRNN \cite{indrnn-skeleton} & 81.8 & 88.0 \\
HCN \cite{li2018co-human-skeleton-ijcai18} & 86.5 & 91.1 \\
\hline
ST-GR \cite{st-graph-routing-skeleton-aaai19} & 86.9 & 92.3 \\
AS-GCN \cite{AS-GCN-skeleton-cvpr19} & 86.8 & 94.2 \\
2s-AGCN \cite{2s-AGCN} & 88.5 & 95.1 \\
AGC-LSTM \cite{attention-gcn-lstm-cvpr2019} & 89.2 & 95.0 \\
DGNN \cite{dgnn} & 89.9 & 96.1 \\
\hline
GR-GCN \cite{GR-GCN} & 87.5 & 94.3 \\
\hline
MS-G3D Net (Joint Only) & 89.4 & 95.0 \\
MS-G3D Net (Bone Only) & 90.1 & 95.3 \\
\textbf{MS-G3D Net} & \textbf{91.5} & \textbf{96.2} \\
\hline
\end{tabular}
\caption{Classification accuracy comparison against state-of-the-art methods on the NTU RGB+D 60 Skeleton dataset.}
\label{tab:ntu-60-results}
\vspace{-0mm}%
\end{table}

\begin{table}[t]
\centering
\begin{tabular}{lcc}
\hline
\multirow{2}{*}{\textbf{Methods}} & \multicolumn{2}{c}{\textbf{Kinetics Skeleton 400}} \\
 & Top-1 (\%) & Top-5 (\%) \\
\hline
\hline
ST-GCN \cite{ST-GCN} & 30.7 & 52.8 \\
AS-GCN \cite{AS-GCN-skeleton-cvpr19} & 34.8 & 56.5 \\
\hline
ST-GR \cite{st-graph-routing-skeleton-aaai19} & 33.6 & 56.1 \\
2s-AGCN \cite{2s-AGCN} & 36.1 & 58.7 \\
DGNN \cite{dgnn} & 36.9 & 59.6 \\
\hline
\textbf{MS-G3D Net} & \textbf{38.0} & \textbf{60.9} \\
\hline
\end{tabular}
\caption{Classification accuracy comparison against state-of-the-art methods on the Kinetics Skeleton 400 dataset.}
\label{tab:kinetics-skeleton-results}
\vspace{-3mm}%
\end{table}

\paragraph{Exploring G3D Configurations.}

Table \ref{tab:g3d-ablations} also compares various G3D settings, including different values of $\tau$, $d$, and the number of G3D pathways in STGC blocks.
We first observe that all configurations consistently outperform the baseline, confirming the stability of MS-G3D as a robust feature extractor.
We also see that $\tau=5$ give slightly better results, but the gain diminishes at $\tau=7$ as the aggregated features become too generic due to the oversized local spatial-temporal neighborhood, thus counteracting the benefits of larger temporal coverage.
The dilation rate $d$ has varying effects:
(1) when $\tau=3$, $d=1$ underperforms $d\in\{2,3\}$, justifying the need for larger temporal contexts;
(2) larger $d$ has marginal benefits, as its larger temporal coverage
come at a cost of temporal resolution (thus \textit{coarsened} skeleton motions).
We thus observe better results when \textit{two} G3D pathways with $d=(1,2)$ are combined, and as expected, we obtain the best results when the temporal resolution is unaltered by setting $\tau= (3,5)$.

\paragraph{Cross-spacetime Connectivity.}

To demonstrate the need for cross-spacetime edges in $\mathcal{G}_{(\tau)}$ defined in Eq.~\ref{eq:g3d-graph-definition} instead of simple, \textit{grid-like} temporal self-edges (on which G3D also applies), we contrast different connectivity schemes in Table \ref{tab:ablation-graph-connectivity} while fixing other parts of the architecture. The first two settings refer to modifying the block adjacency matrix $\tilde{\mathbf{A}}_{(\tau)}$ such that: (1) the blocks $\tilde{\mathbf{A}}$ on the main diagonal are kept, the blocks on superdiagonal/subdiagonal is set to $\mathbf{I}$, and the rest set to $\mathbf{0}$; and (2) all blocks but the main diagonal of $\tilde{\mathbf{A}}$ are set to $\mathbf{I}$. Intuitively, the first produces ``3D grid'' graphs and the second includes extra dense self-edges over $\tau$ frames. Clearly, while all settings allow unified spatial-temporal graph convolutions, cross-spacetime edges as skip connections are essential for efficient information flow.

\paragraph{Joint-Bone Two-Stream Fusion.}

We verify our method under the joint-bone fusion framework on the NTU RGB+D 60 dataset in Table \ref{tab:ntu-60-results}. Similar to \cite{2s-AGCN}, we obtain best performance when joint and bone features are fused, indicating the generalizablity of our method to other input modalities.

\subsection{Comparison against the State-of-the-Art}

We compare our full model (Fig.~\ref{fig:architecture}(a)) to the state-of-the-art in Tables \ref{tab:ntu-120-results}, \ref{tab:ntu-60-results}, and \ref{tab:kinetics-skeleton-results}.
Table~\ref{tab:ntu-120-results} compares non-graph~\cite{st-lstm-trust-gates-skeleton-eccv2016,two-stream-attention-lstm-ntu120,rotclips-ntu120,body-pose-evolution-map-ntu-120} and graph-based methods~\cite{2s-AGCN}.
Table~\ref{tab:ntu-60-results} compares non-graph methods \cite{indrnn-skeleton, li2018co-human-skeleton-ijcai18}, graph-based methods with spatial edges~\cite{st-graph-routing-skeleton-aaai19,AS-GCN-skeleton-cvpr19,2s-AGCN,attention-gcn-lstm-cvpr2019, dgnn} and with spatial-temporal edges~\cite{GR-GCN}.
Table~\ref{tab:kinetics-skeleton-results} compares single-stream~\cite{ST-GCN, AS-GCN-skeleton-cvpr19} and multi-stream~\cite{st-graph-routing-skeleton-aaai19, 2s-AGCN, dgnn} methods. On all three large-scale datasets, our method outperforms all existing methods under all evaluation settings.
Notably, our method is the first to apply a \textit{multi-pathway} design to learn both long-range spatial and temporal dependencies and complex regional spatial-temporal correlations from skeleton sequences, and the results verify the effectiveness of our approach.

\section{Conclusion}

In this work, we present two methods for improving skeleton-based action recognition: a disentangled multi-scale aggregation scheme for graph convolutions that removes redundant dependencies between different neighborhoods, and G3D, a unified spatial-temporal graph convolutional operator that directly models spatial-temporal dependencies from skeleton graph sequences. By coupling these methods, we derive MS-G3D, a powerful feature extractor that captures multi-scale spatial-temporal features previously overlooked by factorized modeling. With experiments on three large-scale datasets, we show that our model outperforms existing methods by a sizable margin.
\\
{\fontsize{7.5}{1} \selectfont \textbf{Acknowledgements}: This work was supported by the Australian Research Council Grant DP200103223.
ZL thanks Weiqing Cao for designing figures.}

\clearpage

{\small
\bibliographystyle{ieee_fullname}
\bibliography{main}
}

\end{document}